\documentclass[11pt]{article}

\usepackage[final]{acl}

\usepackage{times}
\usepackage{latexsym}
\usepackage[T1]{fontenc}
\usepackage[utf8]{inputenc}
\usepackage{microtype}
\usepackage{inconsolata}
\usepackage{graphicx}
\usepackage{booktabs}
\usepackage{amsmath}
\usepackage{array}

\newcommand{\driftunc}{Drift+Uncertainty}
\newcommand{\driftunccol}{\shortstack{Drift+\\[-1pt]Uncertainty}}

\makeatletter
\providecommand{\@LN@col}[1]{}
\makeatother

\title{When Does Uncertainty Measure RAG Hallucination Risk?}

\author{
Rajeel Ahmad Ansari \\
Independent Researcher \\
Delhi, India \\
\texttt{rajeelansari@gmail.com}
\And
Tautik Agrahari \\
Independent Researcher \\
Bengaluru, India \\
\texttt{tautikagrahari@gmail.com}
}

\begin{document}
\maketitle

\begin{abstract}
Uncertainty can measure hallucination risk in retrieval-augmented generation (RAG), but only when retrieved evidence constrains the answer. When evidence is present, unsupported factual claims become less likely; when retrieval coverage is poor, the model can answer from internal knowledge and assign high probability to unsupported claims. We test this across four instruction-tuned models on HotpotQA and SQuAD. After matching faithful and hallucinated answers within word-count bins, length-only classifiers fall near chance, while perplexity reaches 0.81--0.91 AUROC on targeted unsupported edits. Varying retrieval depth shows why the score changes meaning: detection can weaken with only the top-ranked passage, then recover after adding a second passage. A combined uncertainty and semantic-drift score improves calibration, and measurements remain above chance on RAGTruth QA. Results support a conditional view of uncertainty as a useful measure of RAG hallucination risk.
\end{abstract}

\section{Introduction}

Retrieval-augmented generation (RAG) systems \citep{lewis-etal-2020-rag} are meant to ground language-model answers in retrieved evidence. They can still produce answers that sound plausible while making claims not supported by that evidence \citep{ji-etal-2023-survey-hallucination}. Token probabilities offer an appealing measurement: if an answer is poorly supported by the retrieved context, the model may assign lower probability to its tokens. Likelihood measures compatibility with the model's conditioning context, so its connection to faithfulness depends on what that context contains.

This distinction matters because retrieval changes the meaning of uncertainty. When retrieved evidence is adequate, unsupported factual edits should be harder to predict from the context, and token uncertainty can expose that mismatch. When retrieval coverage is poor, the model may answer from internal knowledge. The answer can then be unsupported by the retrieved passage while still receiving high probability under the model. The same low perplexity that looks reassuring under good retrieval can be misleading when the retrieved passage is incomplete.

The central question is therefore when uncertainty has the right evidential meaning for hallucination risk. Experiments use Phi-3-mini-4k-instruct (Phi-3-mini), Gemma-3-4B-IT (Gemma-3-4B), Llama-3.1-8B-Instruct (Llama-3.1-8B), and Qwen2.5-14B-Instruct (Qwen2.5-14B) on HotpotQA and SQuAD \citep{yang-etal-2018-hotpotqa,rajpurkar-etal-2016-squad}. The controlled benchmark first removes a length-based shortcut by matching faithful and hallucinated answers within word-count bins. This selection step preserves answer text and compares examples from both classes at matched lengths. The experiments then test whether token uncertainty remains informative, how retrieval depth changes the measurement, and whether the resulting probabilities are usable at thresholds.

The main testable claim is that uncertainty can measure RAG hallucination risk, while retrieval coverage determines what the measurement means. On controlled answers with targeted unsupported factual edits, perplexity is a strong single-feature discriminator after length control. Under sparse retrieval, the same likelihood score can fail because unsupported answers may still receive high model probability. This makes uncertainty a conditional measurement of evidence support.

\section{Method}

Given a question $q$, retrieved context $C$, and answer $a$, the task is to predict whether $a$ is unsupported by $C$. The representation contains one semantic feature and three uncertainty features.

\textbf{Semantic drift.} The answer and retrieved context are embedded with all-MiniLM-L6-v2, and their L2 distance is computed as:
\begin{equation}
\mathrm{Drift}(a,C)=\lVert E(a)-E(C)\rVert_2 .
\end{equation}
This measures how far the answer is from the retrieved evidence in embedding space.

\textbf{Token uncertainty.} Given the already generated answer, token probabilities are computed under the model conditioned on $(C,q)$, then converted into sequence-level perplexity, mean token probability, and minimum token probability. Perplexity (PPL) is the primary uncertainty feature because it measures average conditional surprise over the answer tokens.

\textbf{Combined score.} The \driftunc{} score is a logistic-regression risk score over standardized semantic drift, perplexity, mean token probability, and minimum token probability. Keeping the scorer low-capacity makes the measured quantities, rather than detector capacity, the object of comparison.

\textbf{Baselines and labels.} Non-uncertainty faithfulness proxies are included by reporting the strongest of BERTScore, NLI, and LettuceDetect as the best baseline in Table~\ref{tab:controlled} \citep{zhang-etal-2020-bertscore,williams-etal-2018-mnli,kovacs-recski-2025-lettucedetect}. The comparison also includes uncertainty variants inspired by recent work on token-level factuality, semantic uncertainty, and memorization signals \citep{fadeeva-etal-2024-fact,farquhar2024detecting,shi2024detecting}. Controlled examples are constructed so that the same question and retrieval pipeline produce either a faithful answer or an answer with targeted factual claims not supported by the retrieved passage. For the targeted edit condition, the model is prompted to make one local unsupported change while preserving a short answer style: changing a factual detail such as a name, date, or number, combining separately retrieved facts into an unsupported relation, or adding a plausible detail absent from the retrieved context. The candidate answer is retained only when the judge labels it unsupported, after which length bin matching is applied. Retrieval-shift labels are assigned by a conservative LLM-judge procedure and checked by a second independently prompted judge; the appendix reports agreement.

\textbf{Length control.} For controlled hallucinations, faithful answers are generated with standard RAG and hallucinated answers contain targeted unsupported factual edits. Faithful and hallucinated answers are then matched within word-count bins. After matching, length-only AUROC is 0.500--0.530 across the eight model--dataset cells, while uncertainty features remain discriminative.

\textbf{Evidence conditions.} The controlled comparison tests whether uncertainty remains informative after deconfounding. The retrieval-shift condition lowers retrieval depth and tests whether the measurement survives degraded evidence. The external condition applies the same features to RAGTruth \citep{niu-etal-2024-ragtruth}, a human-annotated RAG hallucination corpus. Discrimination is measured with AUROC, while score calibration is measured with expected calibration error (ECE) and Brier score \citep{guo-etal-2017-calibration}.

These conditions isolate three different sources of evidence. Length control removes a surface cue. Retrieval shift tests whether uncertainty still reflects support from the retrieved passage. RAGTruth tests whether the same measurements remain above chance outside the controlled construction. A single aggregate AUROC would collapse these questions, although they have different interpretations.

\section{Results}

\subsection{Controlled Hallucinations}

Table~\ref{tab:controlled} summarizes the length-controlled benchmark. Perplexity reaches 0.81--0.91 AUROC in every model--dataset cell and is the strongest single feature. The \driftunc{} score has similar discrimination, with AUROC 0.824--0.911, and exceeds the best non-uncertainty baseline in all eight cells.

The table also shows why length control matters. After matching, length-only AUROC is near chance, while uncertainty remains highly discriminative. The controlled result is therefore not explained by answer length.

The controlled results clarify the role of the combined score. Perplexity is already a strong discriminator; the combined score mainly improves calibration and exposes when semantic drift contributes under retrieval shift. In ablations, sequence-level perplexity is within the bootstrap confidence interval of alternative token aggregators on all eight controlled cells, while Semantic Entropy is weaker on six of eight cells. For short RAG answers, sequence-level perplexity is often enough once answer length is deconfounded.

\begin{table}[t]
\centering
\small
\setlength{\tabcolsep}{3pt}
\resizebox{\columnwidth}{!}{%
\begin{tabular}{llcccc}
\toprule
\textbf{Model} & \textbf{Dataset} & \textbf{Len.} & \textbf{PPL} & \textbf{\driftunccol} & \textbf{Best} \\
 & & \textbf{AUROC} & & & \textbf{base.} \\
\midrule
Phi-3-mini & HotpotQA & 0.502 & 0.811 & \textbf{0.842} & 0.741 \\
Phi-3-mini & SQuAD & 0.504 & \textbf{0.881} & 0.859 & 0.695 \\
Gemma-3-4B & HotpotQA & 0.509 & 0.908 & \textbf{0.911} & 0.839 \\
Gemma-3-4B & SQuAD & 0.519 & \textbf{0.871} & 0.868 & 0.753 \\
Llama-3.1-8B & HotpotQA & 0.517 & \textbf{0.883} & 0.846 & 0.740 \\
Llama-3.1-8B & SQuAD & 0.530 & \textbf{0.885} & 0.869 & 0.748 \\
Qwen2.5-14B & HotpotQA & 0.502 & 0.818 & \textbf{0.824} & 0.762 \\
Qwen2.5-14B & SQuAD & 0.500 & \textbf{0.878} & 0.878 & 0.678 \\
\bottomrule
\end{tabular}
}
\caption{Length control removes the answer-length shortcut, while perplexity and \driftunc{} remain discriminative. PPL denotes sequence-level perplexity; Best baseline is the strongest of BERTScore, NLI, and LettuceDetect.}
\label{tab:controlled}
\end{table}

\subsection{Retrieval Shift}

Controlled examples show how uncertainty behaves when unsupported content is inserted into otherwise context-grounded answers. Retrieval shift tests a different failure mode: the retrieved evidence itself may be incomplete. Retrieval is reduced to $k=1$, so the model receives only the top-ranked retrieved passage, and naturally arising unsupported answers are labeled. Each natural $k=1$ cell contains 100 generated answers; the reversal-prone Gemma-3-4B + HotpotQA and Qwen2.5-14B + SQuAD cells contain 71 and 93 unsupported answers, respectively. In this setting, the likelihood-based score can weaken or reverse. Gemma-3-4B + HotpotQA perplexity drops to 0.285 AUROC (95\% CI 0.180--0.400) and Qwen2.5-14B + SQuAD drops to 0.200 AUROC (95\% CI 0.069--0.398), while semantic drift is strongest in those cells.

\begin{figure*}[t]
\centering
\includegraphics[width=.92\textwidth]{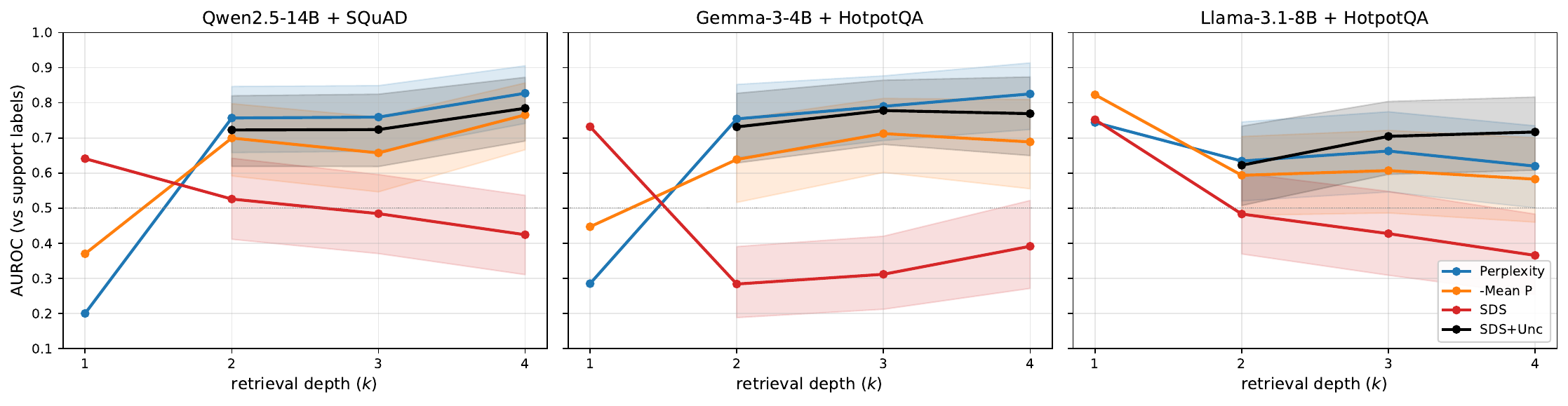}
\caption{Retrieval-depth stress test. In reversal-prone cells, perplexity recovers sharply from one to two retrieved passages, showing that likelihood-based scores depend on retrieval coverage.}
\label{fig:varyk}
\end{figure*}

This pattern is consistent with known out-of-distribution likelihood failures \citep{nalisnick2019do}. High likelihood does not always imply that an answer is faithful to the supplied context. Qualitative inspection shows two recurring failure modes: answers may explicitly abstain while still repeating unsupported entities from the question, or they may complete a plausible answer from parametric knowledge when the top passage is missing the decisive fact. In both cases, the answer can remain easy for the model to score even though it is not supported by the retrieved passage. Figure~\ref{fig:varyk} tests whether the failure is tied to very sparse retrieval; shaded bands show 95\% bootstrap confidence intervals for $k\geq2$, and the appendix reports the corresponding sample sizes. For reversal-prone cells, increasing retrieval from $k=1$ to $k=2$ restores strong perplexity AUROC; later increases to $k=3$ and $k=4$ are much smaller. Across five random seeds, hallucination rate mediates the strength of perplexity (Spearman $\rho=-0.805$), so one retrieval depth can exaggerate or hide the failure.

The retrieval result gives a condition for interpreting likelihood-based scores, not a blanket recommendation to increase $k$. More passages can introduce distractors, and raw hallucination rates do not always improve monotonically. The key point is that likelihood-based scores need enough retrieved evidence for unsupported content to receive lower token probabilities. If an answer can be generated from internal knowledge with low perplexity, that low perplexity is not evidence of faithfulness.

\subsection{External Transfer and Calibration}

On RAGTruth QA, the same measurements remain above chance under a more heterogeneous labeling source. Across 875 human-annotated QA examples, mean perplexity AUROC is 0.584 and mean \driftunc{} AUROC is 0.643. This external validation shows that the controlled result is not an artifact of targeted unsupported edits.

Usable risk measurements need both discrimination and probability calibration. Thresholds have consistent meaning only when scores are calibrated. On the eight controlled cells, mean ECE drops from 0.072 for Platt-scaled perplexity to 0.035 for \driftunc{}, and mean Brier score drops from 0.187 to 0.150. Every controlled cell admits a threshold with at least 95\% precision, with recall concentrated on the clearest unsupported answers. These threshold metrics support high-precision triggers for conservative flagging, retrieval expansion, or human review.

\section{Discussion}

The results support a retrieval-aware interpretation of uncertainty. When retrieval coverage is adequate and the hallucination is an unsupported corruption of context-grounded content, token uncertainty is a strong measurement. When retrieval is sparse, unsupported answers can come from internal knowledge, and likelihood-based scores may fail. The same numerical score can therefore mean different things depending on whether the retrieved passage contains enough evidence to constrain the answer.

This interpretation also separates discrimination from calibration. Perplexity is often the strongest feature in the controlled setting. The combined score matters because it gives better-calibrated probabilities and shows when semantic drift contributes under retrieval or corpus shift. High-precision thresholds can flag answers for rechecking; low-risk cases still depend on the retrieved evidence and the application tolerance for missed errors.

Uncertainty is not a property of the answer alone. It is a property of the answer under a particular retrieved context and scoring model. Length artifacts can make a detector appear better than it is; sparse retrieval can make unsupported answers look probable; poor calibration can make thresholds inconsistent. Separating these effects clarifies when uncertainty measures hallucination risk and when it instead reflects answer length, retrieval coverage, or score calibration.

\section{Conclusion}

Uncertainty measures RAG hallucination risk when the retrieved evidence gives unsupported content lower probability. Length-controlled evaluation shows that perplexity is highly discriminative for targeted unsupported edits, while retrieval-shift experiments show where that success changes meaning. Retrieval coverage determines the meaning of the likelihood score, and calibration determines whether probabilities can be interpreted consistently at thresholds.

\section*{Limitations}

The experiments focus on retrieval-augmented QA rather than long-form writing, tool-using agents, or multimodal generation. The method requires token probabilities, which may be unavailable for some black-box APIs. Controlled hallucinations use targeted unsupported factual edits, and retrieval-stress labels use LLM judges; length control, cross-judge agreement checks, and RAGTruth validation make these choices explicit. RAGTruth provides external validation for QA-style RAG, while broader generation and agent settings require separate evaluation before the same thresholds or failure patterns are assumed to transfer.

\bibliography{custom}

\appendix

\section{Supplementary Analyses}

\subsection{Length-Control Check}

Table~\ref{tab:length-app} reports the matched sample sizes and length-only AUROC values after bin matching. The near-chance values support the claim that controlled results are not driven by answer length.

\begin{table}[h]
\centering
\small
\setlength{\tabcolsep}{4pt}
\begin{tabular}{llccc}
\toprule
\textbf{Model} & \textbf{Dataset} & \textbf{$n$} & \textbf{Non-H} & \textbf{Len. AUROC} \\
\midrule
Phi-3-mini & HotpotQA & 2208 & 25.75 & 0.502 \\
Phi-3-mini & SQuAD & 1882 & 25.50 & 0.504 \\
Gemma-3-4B & HotpotQA & 1414 & 26.79 & 0.509 \\
Gemma-3-4B & SQuAD & 1238 & 27.78 & 0.519 \\
Llama-3.1-8B & HotpotQA & 774 & 30.66 & 0.517 \\
Llama-3.1-8B & SQuAD & 746 & 30.04 & 0.530 \\
Qwen2.5-14B & HotpotQA & 786 & 27.64 & 0.502 \\
Qwen2.5-14B & SQuAD & 836 & 27.06 & 0.500 \\
\bottomrule
\end{tabular}
\caption{Length control after bin matching. The length-only classifier is near chance in every controlled cell. Non-H denotes mean non-hallucinated answer length; matched hallucinated lengths are within the same bins.}
\label{tab:length-app}
\end{table}

\subsection{RAGTruth External Validation}
RAGTruth is used only as an external validation corpus. The fixed responses from the QA split are scored with each of the four scoring models, using the same features as the controlled benchmark. The fixed RAGTruth responses were originally generated by several source models; these source models are not the scoring models.

Table~\ref{tab:ragtruth-app} gives the per-scoring-model external validation results. AUROC is lower than in the controlled benchmark, but the combined score remains above chance for every scoring model.

\begin{table}[h]
\centering
\small
\setlength{\tabcolsep}{4pt}
\begin{tabular}{lcccc}
\toprule
\textbf{Scoring model} & \textbf{PPL} & \textbf{Mean P} & \textbf{Drift} & \textbf{\driftunccol} \\
\midrule
Phi-3-mini & 0.606 & 0.631 & 0.519 & \textbf{0.648} \\
Gemma-3-4B & 0.515 & 0.636 & 0.519 & \textbf{0.650} \\
Llama-3.1-8B & 0.599 & 0.613 & 0.519 & \textbf{0.629} \\
Qwen2.5-14B & 0.615 & 0.646 & 0.519 & \textbf{0.644} \\
\midrule
\textbf{Mean} & 0.584 & 0.632 & 0.519 & \textbf{0.643} \\
\bottomrule
\end{tabular}
\caption{External validation on 875 RAGTruth QA examples. Mean P denotes mean token probability; Drift denotes semantic drift. The combined score remains above chance for every scoring model.}
\label{tab:ragtruth-app}
\end{table}

\subsection{Calibration and Thresholds}

Table~\ref{tab:calibration-app} reports aggregate calibration and high-precision threshold metrics over the eight controlled cells. The combined score improves calibration, while raw perplexity has slightly higher recall at the same precision target.

\begin{table}[h]
\centering
\small
\begin{tabular}{lcc}
\toprule
\textbf{Metric} & \textbf{PPL} & \textbf{\driftunccol} \\
\midrule
Mean ECE & 0.072 & \textbf{0.035} \\
Mean Brier score & 0.187 & \textbf{0.150} \\
Median recall at $\geq$95\% precision & \textbf{0.366} & 0.297 \\
\bottomrule
\end{tabular}
\caption{The combined score improves probability calibration, while raw perplexity has slightly higher recall at the high-precision threshold.}
\label{tab:calibration-app}
\end{table}

\subsection{Retrieval-Sensitivity Analysis}
\begin{figure}[!t]
\centering
\includegraphics[width=\columnwidth]{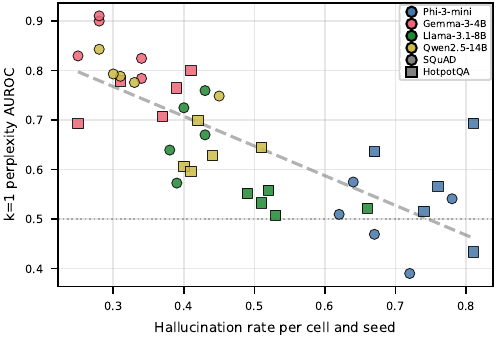}
\caption{Higher unsupported-answer rates correspond to lower $k=1$ perplexity AUROC across random samples (Spearman $\rho=-0.805$). Colors and marker shapes distinguish scoring models and datasets.}
\label{fig:protocol-hallrate}
\end{figure}

Figure~\ref{fig:protocol-hallrate} shows the five-seed sensitivity analysis for the $k=1$ setting. Across random samples, cells with higher unsupported-answer rates tend to have lower perplexity AUROC. Table~\ref{tab:retrieval-counts-app} gives the sample sizes, unsupported-answer counts, and bootstrap intervals for the retrieval-depth stress tests.

\begin{table}[!t]
\centering
\scriptsize
\setlength{\tabcolsep}{2pt}
\begin{tabular}{lrrrr}
\toprule
\textbf{Cell} & \textbf{$k$} & \textbf{$n$} & \textbf{Unsup.} & \textbf{PPL AUROC (95\% CI)} \\
\midrule
\multicolumn{5}{l}{\textit{Natural stress}} \\
Gemma-3-4B + HotpotQA & 1 & 100 & 71 & 0.285 [0.180, 0.400] \\
Qwen2.5-14B + SQuAD & 1 & 100 & 93 & 0.200 [0.069, 0.398] \\
\midrule
\multicolumn{5}{l}{\textit{Seeded vary-$k$}} \\
Qwen2.5-14B + SQuAD & 2 & 100 & 43 & 0.757 [0.658, 0.847] \\
Qwen2.5-14B + SQuAD & 3 & 100 & 43 & 0.759 [0.662, 0.849] \\
Qwen2.5-14B + SQuAD & 4 & 100 & 48 & 0.827 [0.741, 0.905] \\
Gemma-3-4B + HotpotQA & 2 & 100 & 30 & 0.754 [0.644, 0.852] \\
Gemma-3-4B + HotpotQA & 3 & 100 & 32 & 0.790 [0.693, 0.877] \\
Gemma-3-4B + HotpotQA & 4 & 100 & 26 & 0.825 [0.724, 0.914] \\
Llama-3.1-8B + HotpotQA & 2 & 100 & 42 & 0.634 [0.521, 0.746] \\
Llama-3.1-8B + HotpotQA & 3 & 100 & 34 & 0.663 [0.546, 0.775] \\
Llama-3.1-8B + HotpotQA & 4 & 100 & 33 & 0.620 [0.501, 0.735] \\
\bottomrule
\end{tabular}
\caption{Retrieval-stress sample sizes and bootstrap intervals. Natural stress rows report the $k=1$ cells highlighted in the main text; seeded rows correspond to Figure~\ref{fig:varyk}.}
\label{tab:retrieval-counts-app}
\end{table}

\subsection{Label Agreement}
Retrieval-shift labels use a conservative double-judge labeling procedure. A large instruction-tuned LLM labels whether each answer is supported by the retrieved context, and a second independently prompted LLM provides an agreement check. Table~\ref{tab:agreement-app} reports agreement for both the natural $k=1$ stress labels and the three retrieval-depth sensitivity cells used in Figure~\ref{fig:varyk}. Across the natural $k=1$ labels, agreement is 0.998 with Cohen's $\kappa=0.995$; across the vary-$k$ labels, overall agreement is 0.939 with Cohen's $\kappa=0.865$.

\begin{table}[!htbp]
\centering
\small
\setlength{\tabcolsep}{5pt}
\begin{tabular}{lccc}
\toprule
\textbf{Cell} & \textbf{$k$} & \textbf{Agreement} & \textbf{$\kappa$} \\
\midrule
Natural stress cells & 1 & 0.998 & 0.995 \\
Qwen2.5-14B + SQuAD & 2--4 & 0.963 & 0.926 \\
Gemma-3-4B + HotpotQA & 2--4 & 0.913 & 0.798 \\
Llama-3.1-8B + HotpotQA & 2--4 & 0.940 & 0.871 \\
\midrule
\textbf{Overall vary-$k$} & 2--4 & 0.939 & 0.865 \\
\bottomrule
\end{tabular}
\caption{Cross-judge agreement for retrieval-shift labels. The natural stress row pools the five $k=1$ cells used for the natural retrieval-shift analysis.}
\label{tab:agreement-app}
\end{table}

\end{document}